\documentclass[10pt,conference]{IEEEtran}
\IEEEoverridecommandlockouts

\usepackage{amsmath,amssymb,amsfonts}
\usepackage{algorithmic}
\usepackage{graphicx}
\usepackage{textcomp}
\usepackage{xcolor}

\usepackage[utf8]{inputenc} 
\usepackage[T1]{fontenc}
\usepackage{graphicx}
\usepackage{titlesec}
\usepackage{color} 
\usepackage[bookmarksdepth=2]{hyperref} 
\usepackage[misc,geometry]{ifsym}  
\usepackage{float} 
\usepackage[backend=bibtex, style=numeric-comp, sorting=nyt, giveninits=false, maxnames=3, natbib, defernumbers=true]{biblatex}

\usepackage{amsmath,amssymb,amsfonts} 
\usepackage{multirow} 
\usepackage{mdframed} 
\usepackage{xspace} 
\usepackage{soul} 
\usepackage{subcaption} 
\usepackage{enumitem} 
\usepackage[many]{tcolorbox} 
\usepackage{rotating} 
\usepackage{tabularx} 
\usepackage{spverbatim} 
\usepackage{siunitx} 
\usepackage{booktabs} 
\usepackage{listings} 
\usepackage{textcomp} 
\usepackage{makecell}
\usepackage{titlecaps}
\usepackage{tabularx} 
\usepackage{flushend}
\usepackage{balance}

\def\BibTeX{{\rm B\kern-.05em{\sc i\kern-.025em b}\kern-.08em
    T\kern-.1667em\lower.7ex\hbox{E}\kern-.125emX}}

\makeatletter 
\newcommand{\linebreakand}{%
  \end{@IEEEauthorhalign}
  \hfill\mbox{}\par
  \mbox{}\hfill\begin{@IEEEauthorhalign}
}
\makeatother 

\newcommand{\code}[1]{\texttt{#1}}

\newtcolorbox{mybox}[2][]{
    top=0.15in,left=4pt,right=4pt,bottom=4pt,
    fonttitle=\bfseries,
    colbacktitle=gray,
    colback=gray!5,
    colframe=gray!40!black,
    enhanced,
    attach boxed title to top left={xshift=1.5em,yshift=-\tcboxedtitleheight/2},
    boxed title style={size=small},
    drop shadow={black!50!white},
    title=#2,#1
}

\AtEveryBibitem{
  \ifentrytype{article}{
    \clearfield{url}
    \clearfield{doi}
    \clearfield{isbn}
    \clearfield{publisher}
    \clearfield{booktitle}
    \clearfield{pages}
    \clearfield{numpages}
    \clearfield{location}
    \clearfield{issn}
  }{}
  \ifentrytype{inproceedings}{
    \clearfield{url}
    \clearfield{doi}
    \clearfield{isbn}
    \clearfield{publisher}
    \clearfield{pages}
    \clearfield{numpages}
    \clearfield{location}
  }{}
}

\begin{document}
    \title{Real-time Adapting Routing (RAR): Improving Efficiency Through Continuous Learning in Software Powered by Layered Foundation Models}

    \author{
    \IEEEauthorblockN{
        Kirill Vasilevski\IEEEauthorrefmark{1}, 
        Dayi Lin\IEEEauthorrefmark{1}, 
        Ahmed E. Hassan\IEEEauthorrefmark{2}
    }
    \IEEEauthorblockA{\IEEEauthorrefmark{1}Centre for Software Excellence, Huawei Canada}
    \IEEEauthorblockA{\IEEEauthorrefmark{2}Queen's University, Kingston, Canada}

    \vspace{1pt}
    \IEEEauthorblockA{\{kirill.vasilevski, dayi.lin\}@huawei.com, hassan@queensu.ca} 
}

    \maketitle
    \IEEEdisplaynontitleabstractindextext
    \IEEEpeerreviewmaketitle
    \begin{abstract}
To balance the quality and inference cost of a Foundation Model (FM, such as large language models (LLMs)) powered software, people often opt to train a routing model that routes requests to FMs with different sizes and capabilities. Existing routing models rely on learning the optimal routing decision from carefully curated data, require complex computations to be updated, and do not consider the potential evolution of weaker FMs. In this paper, we propose Real-time Adaptive Routing (RAR), an approach to continuously adapt FM routing decisions while using guided in-context learning to enhance the capabilities of weaker FM. The goal is to reduce reliance on stronger, more expensive FMs. We evaluate our approach on different subsets of the popular MMLU benchmark. Over time, our approach routes 50.2\% fewer requests to computationally expensive models while maintaining around 90.5\% of the general response quality. In addition, the guides generated from stronger models have shown intra-domain generalization and led to a better quality of responses compared to an equivalent approach with a standalone weaker FM.
\end{abstract}

    \section{Introduction}
\label{sec:intro}

Due to recent advances in their capabilities, foundational models (FMs) such as large language models (LLMs) have been applied to a wide variety of use cases such as open-ended conversations, planning, code generation, and question answering ~\citep{llm_survey}. 
Developers of FM-powered software (i.e., FMware) \citep{se30} often face a trade-off between maximizing language model capabilities and minimizing the compute resources and costs. Choosing a large FM that has hundreds of billions of parameters will give them better capabilities (e.g. reasoning) and quality of responses when compared to a smaller model that has only a few billion parameters. However, large FMs require magnitudes more expensive computing resources to train and infer \citep{scaling_law}. 
At the same time, small FMs \citep{phi3, llama3.2, apple_intelligence, mobilellm} have recently shown steady improvement in their capabilities, often having adequate performance for common use cases such as text completion, question answering, and instruction following.

To address such a dilemma, it has become increasingly common for FMware developers to opt to combine the usage of large and small FMs as a layered architecture. For requests that a small, weaker FM can handle, the small FM is utilized to save computing costs. When the request is deemed beyond the capability of the small FM, a large FM with stronger capability is used as a fall-back option to guarantee the output quality. Such a strategy can be seen on both cloud-based FMware (e.g., chatbots that use GPT-3.5 by default but fall back to GPT-4 for difficult tasks) and edge-based FMware (e.g., AI assistants on smartphones that use on-device small FM by default but fall back to server-side large FM when needed). For edge-based FMware, such a strategy has added benefits of low internet dependencies, low latency, reduced computational cost due to the use of edge hardware, and enhanced privacy as user data never leaves the device. 

The effectiveness of such a layered architecture depends on the performance of the model routing method. 
A number of solutions for model routing have been proposed in the literature. These can be broadly categorized into using machine-learning-based routers to predict model selection \citep{routellm, hybridllm, frugallm, automix, moe_router, routerbench}, ensembling calls to multiple FMs and selecting the best output \citep{llm-blender, routerbench}, and cascading model inferences until an acceptable response is returned \citep{chen2023frugalgptuselargelanguage}. 
However, many of the above methods have their own set of limitations including redundant inference and latency costs, reliance on training dataset generalization, and complexity of adaption to new data.

\begin{figure}[!tb]
\centerline{\includegraphics[bb=0 0 320 150, scale=0.8]{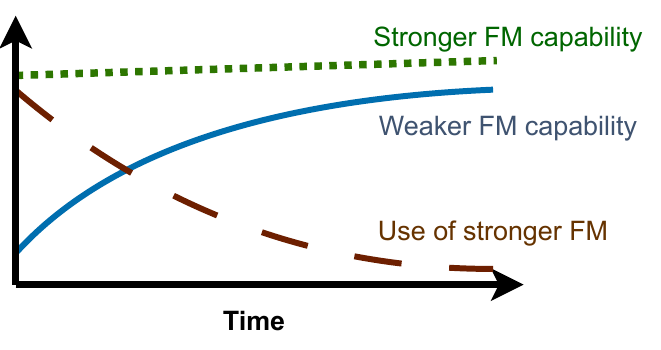}} 
\caption{A demonstration of Real-time Adapting Routing (RAR) objective. Over time, the goal of the system is to reduce reliance (brown, dashed) on the costlier stronger FM by enhancing the capabilities of the cheaper weaker FM (blue, solid) in order to maintain similar levels of capabilities (green, dotted).}
\label{fig:cap_comparison_plateau}
\end{figure}

In this paper, we propose Real-time Adapting Router (RAR), a method that adapts to the evolution of FM capabilities and improves model routing decisions over time, intending to decrease overall computation costs while maintaining the quality of responses. The proposed approach improves upon static model-based routing methods (e.g. ones in RouteLLM \citep{routellm}) by enhancing the weaker FM capabilities with continual learning from the stronger FM as the system is in use, and dynamically adjusting routing decisions to increase utilization of the weaker FM and decreasing overall inference costs (Figure ~\ref{fig:cap_comparison_plateau}). Taking inspiration from continual learning FM-powered agents and in-context learning, our approach uses step-by-step reasoning from the stronger FM as an \textit{in-context instruction guide} (hereinafter referred to as \textit{guide}) to assist the smaller and less capable FM to successfully complete given tasks. 
In our evaluation, RAR achieves a minimum 50.2\% reduction in the usage of stronger FM while maintaining 90\% of output quality when evaluated on several subsets of MMLU \citep{mmlu} benchmark. Furthermore, results demonstrated that the guides generated by the stronger FM generalize across different in-distribution problems (referred to as \textit{intra-domain} generalization), showing that our approach is not simply memorizing individual solutions but instead allows useful knowledge to be re-used.

The remainder of the paper is structured as follows. Section~\ref{sec:related_work} provides an overview of different methods to model routing and layering, followed by an overview of chain-of-though prompting and reasoning generation, as well as their applications within the context of continual learning.
Section~\ref{sec:routeadapt} then introduces our proposed approach. 
Section~\ref{sec:evalaution} presents our experimental setup as well as evaluation results of RAR on a subsets of MMLU question-answering dataset.
Section~\ref{sec:threats} describes several threats to the validity of our study.
Lastly, Section~\ref{sec:conclusion} concludes the paper.
    \section{Background and Related Work}
\label{sec:related_work}

In this section, we provide background information on underlying concepts relevant to RAR. We describe existing methods for routing and layering FM inference in Section ~\ref{sec:related_work:routers}, prompt-based FM enhancement techniques in Section ~\ref{sec:related_work:llm_enh}, and introduce continual learning concepts and their applications with FMs in Section ~\ref{sec:related_work:cl}.

\subsection{Model Routing and Layering}
\label{sec:related_work:routers}

As introduced in Section~\ref{sec:intro}, a model routing method aims to balance the quality of FM-generated output and the associated inference costs by selecting the most optimal model for a given request. Two major categories of routing methods are predictive and non-predictive routing \citep{routerbench}. Non-predictive routing is based on collecting FM-generated outputs, typically done sequentially, until an answer passes some quality threshold \citep{chen2023frugalgptuselargelanguage, tabi}, which is one of its limiting factors given that there is an increased cost due to many rounds of model inference. On the other hand, predictive routing uses the contents of the input request to predict the optimal model selection, bypassing the need for FM output and thus leading to reduced costs and latency. Predictive routing can be implemented by training machine learning models on supervised classification \citep{sl_route} or ranking tasks \citep{routellm, llm-blender} using a dataset of input requests and associated model preference labels \citep{routerbench, routellm}. As an example, RouteLLM \citep{routellm} uses a prompt-model human preference dataset \cite{mtbench} to train several ranking and classification models to predict the optimal model selection, and demonstrates notable reduction in costs without significant compromise in the quality of responses. However, the performance of predictive methods is often limited by the quality of the training dataset and how well it can represent the real-world distribution of user inputs. As shown in \citep{routellm}, a router trained only on a human preference dataset demonstrates similar performance to a random baseline when evaluated on two different problem-solving benchmarks, highlighting the impact of careful training data selection on performance. Additionally, the capabilities of many model-based routers are \textit{static} post-deployment and require re-training and re-deployment whenever the training dataset or FM capabilities are updated.

Model routing can be applied to both cloud-based and edge-based FMware. Cloud-based chat applications such as OpenAI's ChatGPT \citep{chatgpt} and Anthropic's Claude \citep{claude} can benefit from routing methods by sending simpler requests to less compute-intensive versions of their underlying FMs, and more complex requests to larger models, all without sacrificing the quality of response. Edge-based FMware often utilizes \textit{cloud-edge collaboration} - simpler requests could be sent to an edge-hosted FM, meanwhile complex ones are forwarded to a cloud-hosted FM instead. This setup allows to reduce compute costs by offloading some of the computation to the hardware on the edge device, while also preserving the privacy of users as a portion of the data never leaves the device. 

\subsection{Chain-of-Thought and Reasoning}
\label{sec:related_work:llm_enh}

Chain-of-thought (CoT) is a method that asks the FM to explicitly output its reasoning \citep{cot} and has been shown to significantly improve the quality of the generated output, with the added benefit of providing the user with an explicit record of how the model arrived at its answer. Similar trends have been observed in other methods that target reasoning generation \citep{tot, got} to further improve FM performance. Taking note of this, several studies have attempted to use reasoning with in-context learning, where reasoning is a part of the input prompt, to guide FM generation to a desired output \citep{voyager,clin, skills_llm}. Furthermore, re-using step-by-step reasoning outputs as part of the prompt design has been used to guide FMs to perform similar but yet unseen tasks \citep{skills_llm}. This approach can be viewed as an equivalent of \textit{continual learning} paradigm used in conventional machine learning which is described in the next section.


\subsection{Continual Learning}
\label{sec:related_work:cl}

Continual learning (CL), also referred to as \textit{lifelong learning}, is an approach in machine learning that aims to incrementally train models over a lifetime on a dynamic data distribution, compared to conventional methods that are built by learning a static distribution \citep{cl_survey}. CL methods can be categorized into replay (data) based, regularization-based, optimization-based, representation-based, and architecture-based depending on which part of the machine learning pipeline they target \citep{cl_survey}. Nonetheless, most continual learning methods operate by changing the way the machine learning model learns, that is, how model parameters are updated compared to conventional training procedures. 

Prior work has also leveraged in-context learning as a means for continual learning without updating model parameters. 
For example, Voyager \citep{voyager} is an FM-powered lifelong learning agent that autonomously learns how to play the popular video game Minecraft. The agent represents its skill library as a collection of natural language descriptions of how to perform a certain task, which the underlying FM can then use to guide the generation of the output response. Similarly, a CLIN agent~\citep{clin} (stands for "\textbf{c}ontinually \textbf{l}earning from \textbf{in}teractions") aims to utilize multiple trials to continually improve its capabilities across varying tasks and environments. It uses a memory of causal abstractions in natural language to learn useful knowledge that generalizes across trials and environments. The success of in-context continual learning methods served as inspiration for the core functionality of RAR.
    \section{RAR: Real-time Adaptive Routing}
\label{sec:routeadapt}

In this section, we provide an overview of RAR and its different components. The goal of RAR is two-fold: in a layered architecture with a stronger FM and a weaker FM, over time, 1) maintain as closely as possible the overall capability levels of the stronger FM, and 2) reduce the use of strong FM by maximizing the usage and capabilities of the weaker FM (Figure ~\ref{fig:cap_comparison_plateau}). This is achieved by the weaker FM utilizing the \textit{guides} generated by stronger FM, as part of its context to assist in generating a response. 

\begin{figure*}[!tb]
\centerline{\includegraphics[bb=-50 0 800 275,width=\textwidth]{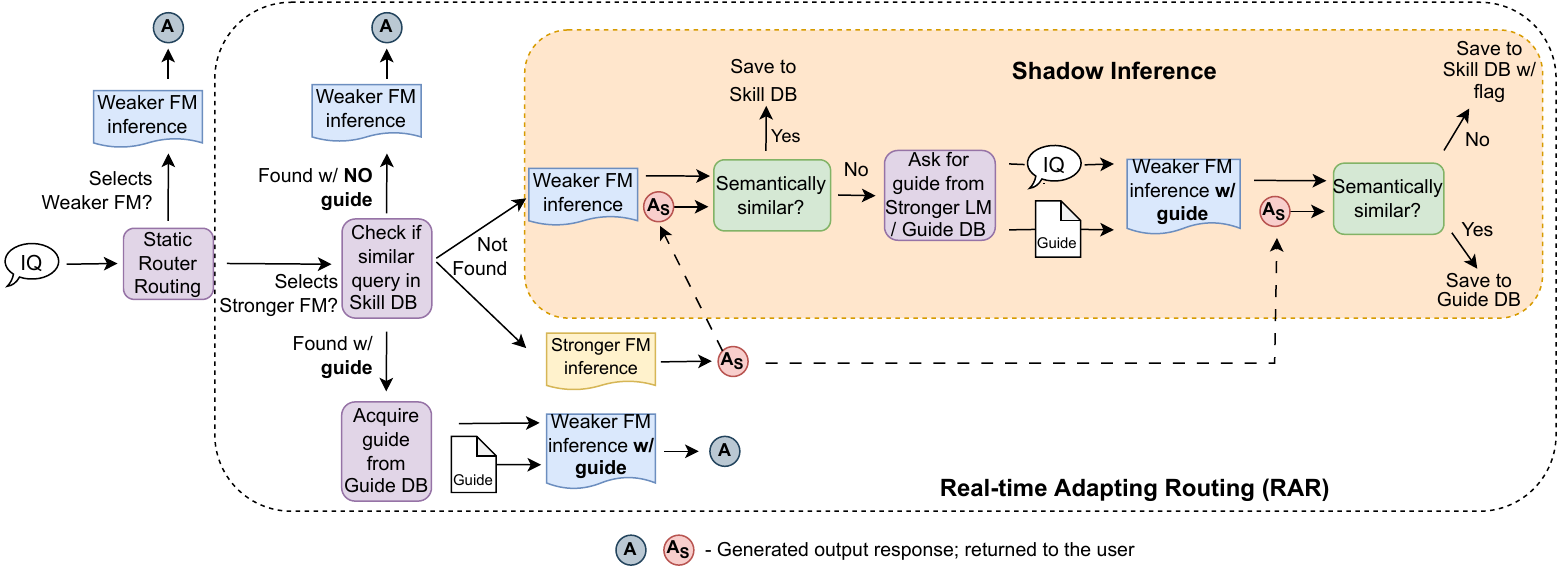}} 
\caption{Overview of RAR procedure and shadow inference, described in Section ~\ref{sec:routeadapt}; IQ - incoming request}
\label{fig:seq_diag}
\end{figure*}

When a user request is received, it is first given to a static router such as the ones described in Section ~\ref{sec:related_work:routers} to obtain a routing decision. In the case that the weaker (e.g. on-device) FM is selected, RAR forwards the request straight to the weaker model since the goal of our method is to use the least compute-intensive model. In the case that the routing decision selects the stronger FM, RAR performs \textit{shadow inference} (described in Section ~\ref{subsec:shadow_inf}) to evaluate whether the weaker FM could still successfully serve the given request, either by itself or with a \textit{guide} provided by a stronger FM. The method is based on the hypothesis that \textbf{the stronger FM can provide insightful and knowledgeable information in the generated guide which can be utilized by weaker FM through in-context learning to improve the quality of the generated response.} Any time that a weaker FM generates an aligned response, the request and guide (if used) are recorded into a skill and guide memory. Future incoming requests are then compared against the ones stored in skill and guide memory which determines whether the new request is sent to the weaker FM and if it requires a guide. Over time, RAR will have a significant collection of useful guides that weaker FM can make use of to successfully serve similar requests, which allows the system to route more samples to the weaker FM rather than the stronger FM as decided by the static router. Below, we describe in detail several components that make up RAR. 

For the cloud-edge collaboration use case, RAR has the added benefit of caching generated guides on the edge device, reducing the need for repeated inference on the expensive stronger FM. Additionally, depending on the the use habits of the user, the weaker FM hosted on the edge also becomes more personalized to the user's needs as the system acquires more guides from the user's requests. The side benefit of enhanced personalization leads to improved user experience as the system can better match user's expectations.

\subsection{System Objectives}

 Given that in a real-world deployment there are little to no guarantees to the domain constraints of the requests, automatically determining the validity of the generated response (without external input from an expert rater, e.g. a user who knows what the response should be) is a very difficult task. As such, when operating in an open-domain environment, the goal becomes not to evaluate the correctness of the system to the unavailable ground truth, but instead to compare how well RAR can maintain the performance of the system close to that of the stronger FM. It is important to note that by \textit{aligned response}, we define the case as a weaker FM generates a \textit{semantically-similar} response to that of a stronger FM, which is different from generating the \textit{correct} response given a request. Furthermore, the output of a RAR method can only be as good as the stronger FM's outputs as it only attempts to mimic stronger FM's capabilities rather than surpass them. Evaluating how similar two outputs are is not a trivial task, and we outline several methods of comparison in the following section.

\subsection{Semantic Comparison of Requests and Responses}

To measure whether two requests or two responses are similar (e.g the weaker FM's response and stronger FM's response), we make use of vector similarity metrics (e.g. cosine or dot product), or LLM-as-a-judge \citep{llm_judge} approach. With the vector similarity method, a user can select a similarity score threshold that delineates whether two requests or two responses are considered similar or not. For the LLM-as-a-judge approach, we ask an FM to compare two requests or two responses and return a single-word answer whether the requests or responses are semantically similar or different. Regardless of the method used, the semantic comparison becomes a binary decision that is used to control the next steps of the proposed method.

\subsection{Initial Static Routing}

In a real-world deployment, RAR would be used in conjunction with an initial static model-based predictive routing approach which has been pretrained to select the correct model based on the request (e.g. any of the ones in RouteLLM \citep{routellm}). As the first step in the routing process, using a model-based router allows to avoid unnecessary model inference by making the first selection of the assumed appropriate model. 

Given that one of the target objectives of RAR is to minimize expensive model inference calls from the stronger FM, any requests that have been forwarded to the weaker FM continue on to model inference without any modifications. The following steps are only involved when the static router selects the stronger FM for inference, that is, the weaker FM is unlikely to successfully serve the request according to the static router.

\subsection{Shadow Inference and Adaptation}
\label{subsec:shadow_inf}

When the static router selects the stronger FM for inference, RAR performs so-called \textit{"shadow inference"} consisting of several operations (Figure ~\ref{fig:seq_diag}). To maintain a good user experience, the request is first forwarded to the stronger FM and the response is then returned to the user. In the background, we then generate a response to the same request with the standalone weaker FM which leads to three distinct cases depending on the response. If the two responses are similar, we refer to this as \textit{Case 1} described further below. If the two responses are different, \textit{Case 2} applies where the system then attempts to improve weaker FM's response with a guide. Lastly, if the weaker FM does not generate an aligned response with guide help, this is considered as \textit{Case 3} where such request is automatically routed to stronger FM in the future and shadow inference is repeated at a later time.

\subsubsection{Case 1: Standalone Weaker FM}

If the weaker FM generates an aligned response to the request without any assistance, the request is embedded into a latent vector (described in detail in Section ~\ref{sec:evaluation:exp_setup}) and is saved into skill memory (Section ~\ref{subsection:memory}). When any future requests come in, they will be compared to existing ones in the skill memory using a similarity score. If the similarity score passes a set threshold, the incoming request is directly forwarded to weaker FM for inference in the future.

\subsubsection{Case 2: Weaker FM with Guide}

Once it is confirmed that the weaker FM is unable to generate an aligned response by itself, the system then evaluates whether the weaker FM can generate an aligned response if a guide is provided. The guide is acquired either from a guide memory (Section ~\ref{subsection:memory}) or generated by the stronger FM. The guide is then used in conjunction with the original request to generate a response using the weaker FM. If the weaker FM generates an aligned response (semantically similar to the stronger FM's response) it is considered that similar samples can also be successfully served by the weaker FM once a guide is provided. As such, the request and the corresponding guide are saved into the guide memory for future use.

\subsubsection{Case 3: Weaker FM Fails With Guide}

If the request is too dissimilar to any of the requests stored in the guide memory, or the acquired guide does not lead to an aligned response, a new guide is requested from the stronger FM and is evaluated for success. If none of the attempted guides lead to an aligned response, the request is saved in skill memory, with a flag that indicates that any future highly similar or identical requests to be routed to stronger FM by default. After a certain period of time (a tuned hyperparameter), any closely similar requests repeat the shadow inference process to assess whether any new guides that have been saved in memory can now lead to an aligned response with the weaker model.

\subsection{Guides}

Guides are generated as instructions or hints that can assist in answering a given request but that do not contain the actual answer. When the system is initially deployed, the guide memory will be empty and the majority of guides are generated by the stronger FM. Over time, the guide memory is populated as more guides are generated and evaluated, leading to new requests being able to reuse existing guides for response generation. The ability to re-use guides across different requests is the desired generalization behavior that differs RAR from simply memorizing solutions where each guide is only relevant to a unique request.

\subsection{Skill and Guide Memory}
\label{subsection:memory}

Skill and guide memory are represented as a vector database that stores the embedding vector of the request, along with the corresponding guide in plain text. Indexing is done by comparison of the embedding vector of the candidate request and existing requests in the database, measured by a similarity score (detailed in Section ~\ref{sec:evaluation:exp_setup}). By varying the similarity score threshold hyperparameter, it is possible to control exploration (generate specific guides with stronger FM) vs. exploitation (use guides from less similar requests) in terms of acquiring a guide. This threshold ranges from zero to one, with a higher threshold meaning the requests must be more similar. Requests that do not require a guide for generating an aligned response are stored without a guide attached compared to those that require a guide, meaning that if the request is similar to an entry that does not contain a guide, this request is considered as \textit{Case 1} or \textit{Case 3} and can be forwarded directly to the corresponding FM for inference. Note that skill and guide memory can be implemented in various ways that will not affect the overall operation of the system and that the approach used in this paper is only one such implementation.

    \section{Evaluation}
\label{sec:evalaution}

In this section, we present the details of an evaluation of applying RAR on a subset of popular MMLU \citep{mmlu} benchmark, along with a comparison to other related methods in FM deployment.

\subsection{Methodology}
\label{sec:evaluation:exp_setup}

\subsubsection{Datasets}
\label{sec:meth:datasets}
To validate the proposed method, we opted for a dataset with a constrained input and defined ground truth to simplify the evaluation of aligned responses. We utilize a subset of MMLU \citep{mmlu} question answering benchmark that consists of multiple-choice questions and answers on various topics. Commonly used for assessing the capabilities of FMs, it gives one measure of differences between FMs which helps select weaker and stronger FMs that have confirmed different capability levels. Questions from MMLU also represent a type of task where the design of the overall input prompt (e.g. using a Chain-of-Thought \citep{cot} prompt) could have a beneficial impact on the correctness of the generated response, which assists in evaluating the usefulness of the generated guides as well as inter-task generalization. We used a version provided by \citep{routerbench} which provides solutions and evaluations to every sample in the benchmark across a variety of FMs including the ones we selected. We utilized a subset of samples our weaker FM failed to successfully answer to explicitly focus on requests that are confirmed more difficult for the weaker FM and assess whether guided generation can be beneficial (Figure ~\ref{fig:stage_diag}). Additionally, we selected the top three domains that contained the largest number of failed samples resulting in a subset of 754 samples of questions on professional law, 359 on high school psychology, and 675 on moral scenarios. 

\subsubsection{Models}
For the weaker FM we selected \code{Mistral-7B-instruct} \citep{mistral7b}, with \code{gpt-4o-2024-08-06} \citep{gpt4o} and \code{Llama-3-70B-instruct} \citep{llama3} as the stronger FMs. We utilized two different stronger FMs in order to ensure that the method behavior is not unique to one specific model. For all latent embeddings, we used the popular \code{all-MiniLM-L12-v2} model \citep{minilm} with 384-dimensional embedding and used cosine similarity for skill and guide memory indexing. Skill and guide memory is implemented using Qdrant DB \citep{qdrantQdrantVector} vector database. To select the similarity score threshold for querying the skill and guide memory, we first measured the median sample-wise cosine similarity score for the MMLU professional law subset, which was 0.442. We then set the similarity score threshold cutoff at 0.2 to encourage more attempts at generation and re-use of existing guides to evaluate guide generalization. However, regardless of the threshold value only the highest-scoring request result is selected in the end. 

\subsubsection{Experiment procedure}
\label{sec:method:exp_proc}
\begin{figure*}[!tb]
\centerline{\includegraphics[bb=-50 0 575 150,width=\textwidth]{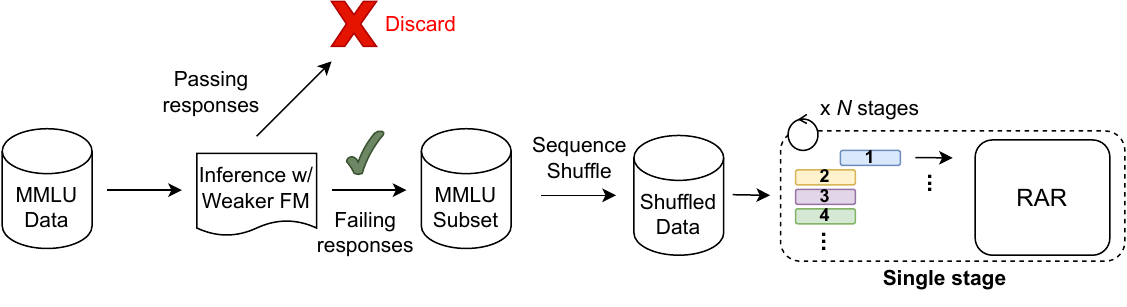}} 
\caption{Overview of data selection and single stage in the experiment process. First, samples from select MMLU domains are sent for inference by the weaker FM. Responses are compared against MMLU ground truth solutions, and the samples that match ground truth are discarded. The failing samples then make up the MMLU subsets described in Section ~\ref{sec:meth:datasets}. Prior to the experiment, the subsets get randomly shuffled to randomize the sample sequence. Then, the samples are used one-by-one by our method in a single stage, with multiple stages per entire experiment.}
\label{fig:stage_diag}
\end{figure*}

To evaluate our proposed method, we attempt to mimic real-world use of the system. In a single \textit{stage}, RAR goes through a given dataset (described in Section ~\ref{sec:meth:datasets}) on a sequential sample-by-sample basis and generates responses. We evaluate the \textit{capability} of the method by how many responses are aligned with the stronger FMs response, and as such, we record the number of aligned responses and the number of times stronger FM is used for every stage (we utilize Chi-square test with 95\% confidence interval to evaluate significance \citep{chi}). The entire experiment consists of multiple stages where the above process is repeated multiple times to simulate repeated inference of similar requests and allow for the RAR method to populate its guide memory. Given that the sequence in which samples are seen impacts the performance of RAR, we randomly shuffle the datasets five times to reduce dependence on the sample sequence and perform the experiment on every permutation. An overview of a single stage is provided in Figure ~\ref{fig:stage_diag}.

\subsection{RQ1: Is it possible to reduce reliance on stronger FM meanwhile maintaining similar capability levels?}
\label{sec:rq1}

\subsubsection{Approach}
  We compared our approach to the following methods: standalone stronger FM, standalone weaker FM, weaker FM with zero-shot CoT \citep{cot} reasoning, and an oracle static router system. Chain-of-Thought approach is included since the use of reasoning in prompts has been shown to improve the quality of generated response \citep{cot}, as well as it can be considered an equivalent approach to the guided generation in RAR for the exception that the latter uses reasoning generated by the stronger FM. The static router represents an ideal predictive \textit{oracle static router} model where the weaker FM only receives samples that it can confidently generate an aligned response to, meanwhile, all the rest are sent to the stronger FM. It is simulated by initially profiling the dataset with the weaker model, then selecting the common subset of aligned requests with the rest being sent to the stronger FM. Unless stated otherwise, results use the best performing static routing setup using \code{gpt-4o-2024-08-06} as the stronger FM.

\subsubsection{Results} 
\begin{figure}[!tb]
\centerline{\includegraphics[bb=0 0 475 310,width=\columnwidth]{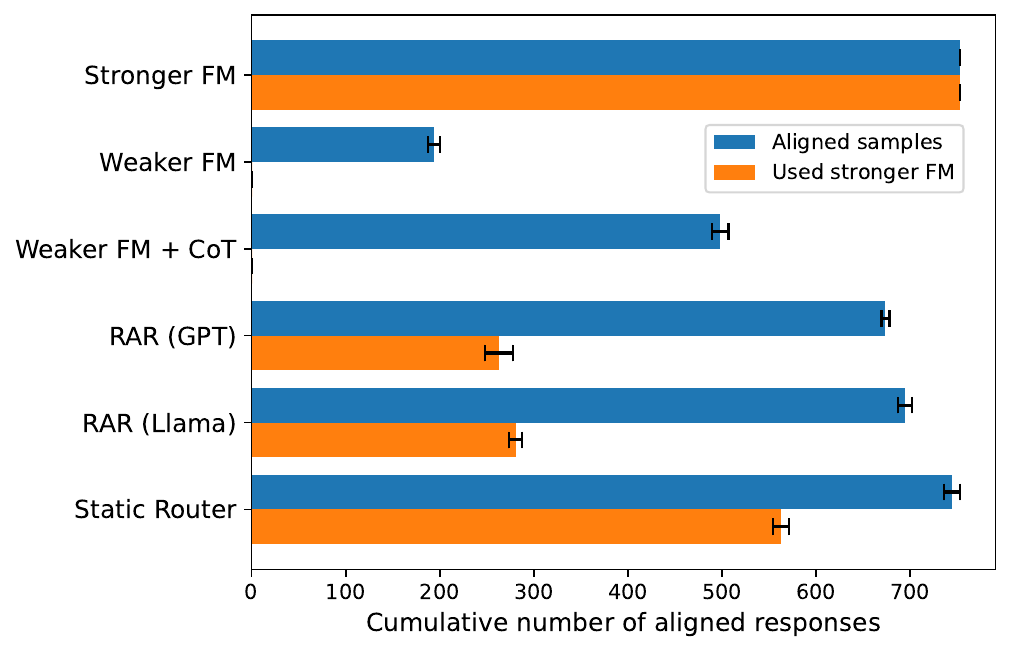}} 
\caption{Cumulative number of aligned responses and calls to the stronger FM on a subset of \textit{MMLU Professional Law} dataset; mean and standard deviation across five random shuffles. RAR \textbf{(GPT)} utilizes \code{gpt-4o-2024-08-06} as the stronger model, and RAR \textbf{(Llama)} uses \code{Llama-3-70B-instruct}; weaker FM is \code{Mistral-7B}. Number of aligned responses in blue (higher is better), use of stronger FM in orange (lower is better)}
\label{fig:mmlu_prof_fig}
\end{figure}

\begin{figure}[!htb]
\centerline{\includegraphics[bb=0 0 455 310,width=\columnwidth]{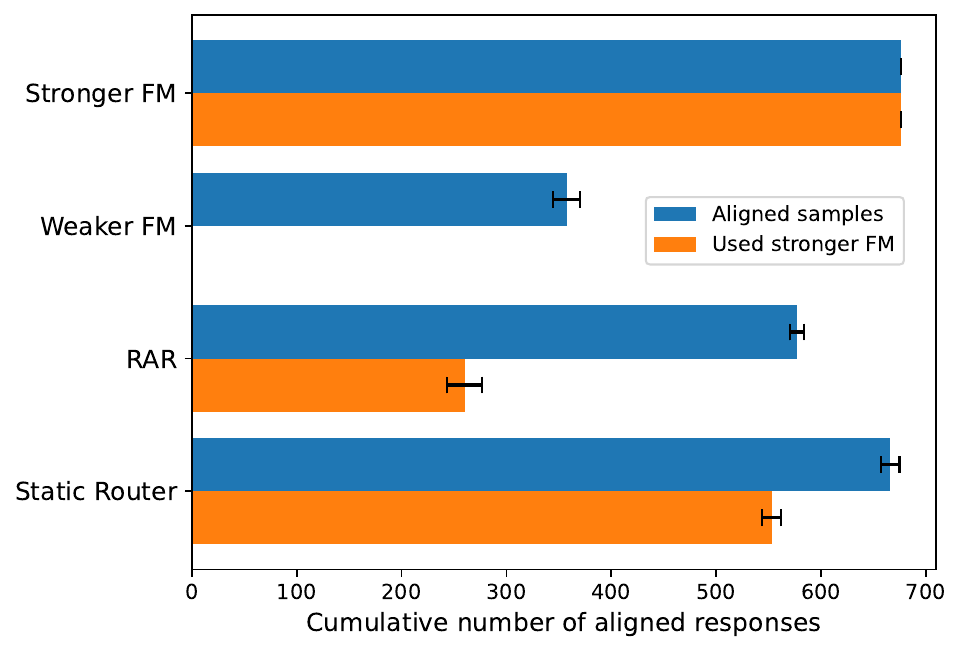}} 
\caption{Cumulative number of aligned responses (blue, higher is better) and calls to the stronger FM (orange, lower is better) on a subset of \textit{MMLU Moral Scenarios} dataset; mean and standard deviation across five random shuffles. RAR utilizes \code{gpt-4o-2024-08-06} as the stronger model; weaker FM is \code{Mistral-7B}.}
\label{fig:mmlu_moral_scen}
\end{figure}

\begin{figure}[!tb]
\centerline{\includegraphics[bb=0 0 455 310,width=\columnwidth]{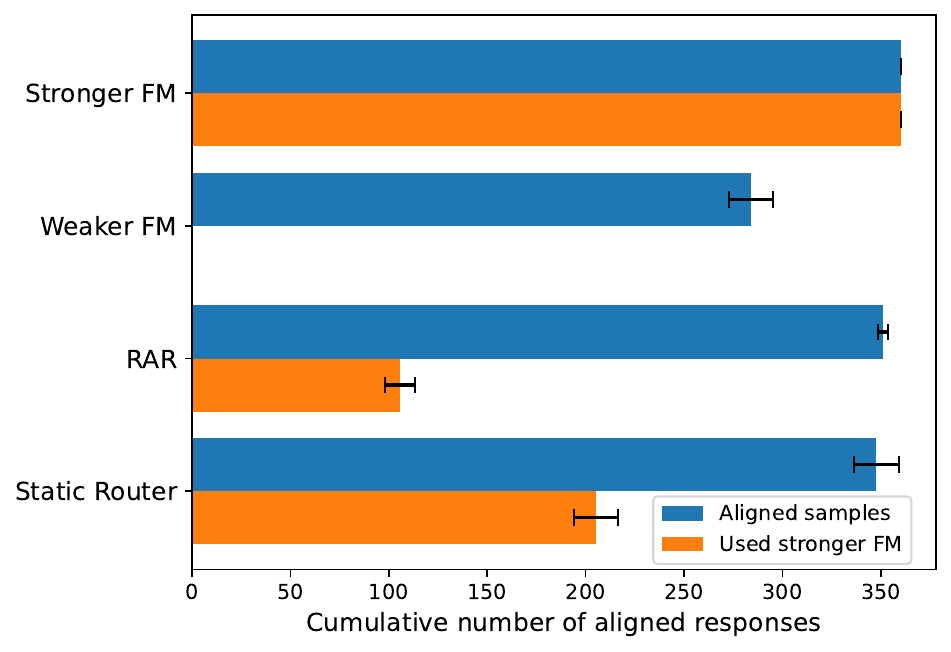}} 
\caption{Cumulative number of aligned responses (blue, higher is better) and calls to the stronger FM (orange, lower is better) on a subset of \textit{MMLU High-School Psychology} dataset; mean and standard deviation across five random shuffles. The first stage is used as a profiling step where standalone weaker FM (no guide) is used to determine failing samples. RAR utilizes \code{gpt-4o-2024-08-06} as the stronger model; weaker FM is \code{Mistral-7B}.}
\label{fig:mmlu_hs_psych}
\end{figure}

\textbf{The proposed method shows a significant reduction of 50.2\% in reliance on the stronger FM while maintaining 90.5\% of response quality compared to oracle static router.} In Figure ~\ref{fig:mmlu_prof_fig}, our approach demonstrates improvements in terms of the number of aligned responses of at least 349\% over weaker FM, 135\% over weaker FM + CoT, and maintaining 90.5\% and 89.5\% of performance compared to oracle static router and standalone stronger FM respectively (p < 0.001 for all). At the same time, our method reduces the use of stronger FM by at least 62.8\% over standalone stronger FM and 50.2\% over the oracle static router respectively (p < 0.001 for all). Similar trends are observed where the task domain is changed to moral scenario questions (Figure ~\ref{fig:mmlu_moral_scen}) and high-school psychology questions (Figure ~\ref{fig:mmlu_hs_psych}), as well as between different choices of stronger FMs (\code{gpt-4o-2024-08-06} vs \code{Llama-3-70B-instruct}).

Simple inclusion of reasoning with CoT approach shows considerable improvement in aligned response generation over standalone weaker FM, and is also confirmed by the performance of the RAR methods. Improved performance of RAR methods over CoT approach demonstrates the benefits of using reasoning generated by the stronger FM compared to the weaker FM, underlining our hypothesis about the increased usefulness of the knowledge provided by the stronger FM when generating the guide compared to reasoning from the weaker FM. 

Compared to the oracle static router, RAR maintains 90.5\% of the quality of responses and reduces the use of the stronger FM by at least 50.1\% (depending on the choice of stronger FM). One interesting finding is the difference in the performance of standalone weaker FM and the static router. The oracle router relies much more on stronger FM (562 mean number of calls) compared to the number of questions for which standalone weaker FM can generate aligned responses (mean of 193). This discrepancy is due to the fact that the standalone weaker FM is able to solve more requests as \textit{all} of the seen samples are attempted 5 times (once per stage, described in Section ~\ref{sec:method:exp_proc}) over the entire duration of the experiment, and thus there is a higher chance of a sample generating an aligned response. On the other hand with the static router, the weaker FM only sees a limited number of samples over the duration of the experiment (determined from profiling), with the remainder being sent to the stronger FM. This is equivalent to a pre-trained static model-based router where even if a new incoming request could be theoretically served by the weaker FM, the router still forwards it to the stronger FM based on the previously learned routing decisions.

\begin{mybox}{Summary}
    \textit{On subsets of MMLU dataset, RAR approach is able to reduce reliance on the stronger FM by 50.1\% while maintaining 90.5\% of the quality of generated responses compared to an oracle static router (p < 0.001). Experimental results demonstrate the benefits of our approach in reducing the costs of FM deployment while maintaining similar levels of capabilities.}
\end{mybox}

\subsection{RQ2: Do the guides provided by the stronger FM exhibit inter- and intra-domain generalization?}

\subsubsection{Approach}
To determine whether the guides collected by RAR contain generalized knowledge, we performed two experiments. The first experiment targets intra-domain generalization in order to understand whether a guide from one question can be beneficial to another related question of the same domain (e.g. professional law). To evaluate, we compare the per-stage number of aligned guided samples based on whether the guide was acquired from the guide memory or freshly generated by the stronger FM. The expectation is that the longer the system operates, the more requests for guides will be fulfilled by the guide memory rather than a new generation from stronger FM, thus, demonstrating intra-domain generalization. 

The second experiment evaluates inter-domain generalization to understand whether guides generated for one domain can be beneficial for questions in another domain. For this, we re-run the same experiment setup as RQ1 in Section ~\ref{sec:rq1} but with two differences: 1) the guide memory is fully populated from the beginning with guides from professional law dataset from RQ1, and 2) we apply our approach on samples from high-school psychology and moral scenarios datasets. During guided generation, the system is only allowed to re-use guides intended for the professional law domain (no new guides allowed) which is achieved by setting the similarity score threshold to a very low arbitrary value (we used 0.1). The expectation is that there will no major increase in the number of aligned responses with the out-of-domain guides when compared to standalone weaker FM generation because the guide prompts are generated to be domain-specific.

\subsubsection{Results} 
\begin{figure}[!tb]
\centerline{\includegraphics[bb=0 0 420 310,width=\columnwidth]{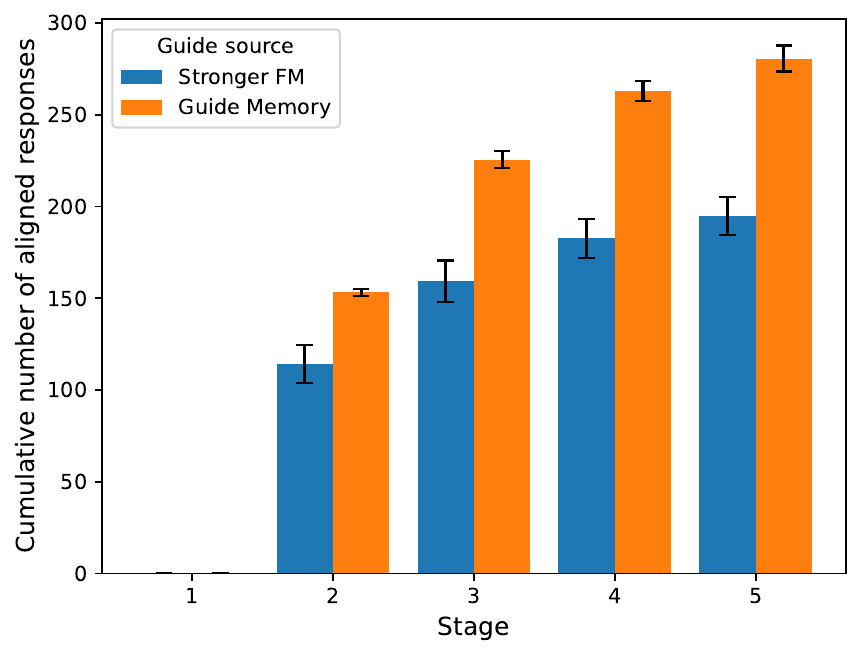}} 
\caption{Cumulative number of aligned guided responses per stage on a subset of \textit{MMLU Professional Law} dataset; guides acquired from stronger FM in blue, and from guide memory in orange; mean and standard deviation across five random shuffles. RAR utilizes \code{Llama-3-70B-instruct} as the stronger model; weaker FM is \code{Mistral-7B}.}
\label{fig:guide_gen_prof_law}
\end{figure}

\textbf{Guides generated by RAR demonstrate intra-domain generalization and even show potential for inter-domain generalization.} Examining results of the first experiment in Figure ~\ref{fig:guide_gen_prof_law}, the difference between the guide generation with stronger FM and the use of guide memory is 34.2\%, 41.6\%, 44.0\%, and 44.4\% for stages 2 to 4 respectively, demonstrating an increase in the use of guide memory as the system acquires more guides. This demonstrates the desired intra-domain generalization behavior where some requests successfully re-use previously generated guides from other requests, leading to reduced costs due to decreased use of the stronger FM. Additionally, the use of either source of guides slowly plateaus as the number of stages increases due to reaching the maximum number of samples that the weaker FM can generate an aligned response to (with or without guide help). Note that the first stage does not use a guide as it is used to identify samples that can be solved solely by the standalone weaker FM. 

Results for the second experiment in Table ~\ref{tab:generalization} demonstrate an interesting dynamic. In both cases where requests from high-school psychology or moral scenarios make use of guides from the professional law domain, the number of aligned responses increases by 6-7\% over the standalone weaker FM approach even though the requests and guides come from very different domains. At the same time, intra-domain guides (e.g. HSP-HSP and MS-MS target-source tasks) show major benefits to aligned response generation with 18.2\% and 32.5\% improvement for high-school psychology and moral scenarios domains respectively. Overall, the inclusion of any type of reasoning as part of the input prompt led to a higher number of aligned responses, which is consistent with findings from previous studies \citep{cot, tot}.

\begin{table}[!tb]
\caption{Inter- and intra-domain guide generalization performance as the difference between the cumulative number of aligned responses vs. stronger FM (lower is better). \textit{HSP} - high-school psychology, \textit{MS} - moral scenarios, \textit{PL} - professional law. Guide source task indicates the source domain used for guide generation; 5-shot inference per sample. Utilizes \code{gpt-4o-2024-08-06} and \code{Mistral-7B} as stronger and weaker FM respectively.}
\begin{center}
\begin{tabular}{@{}llr@{}}
\toprule
Target Task & Guide Source Task & \multicolumn{1}{l}{Difference from Stronger FM} \\ \midrule
HSP         & PL                & 15.0\%                                          \\ \cmidrule(l){2-3} 
            & HSP               & 2.5\%                                           \\ \cmidrule(l){2-3} 
            & Unguided          & 21.1\%                                          \\ \midrule
MS          & PL                & 40.1\%                                          \\ \cmidrule(l){2-3} 
            & MS                & 14.6\%                                          \\ \cmidrule(l){2-3} 
            & Unguided          & 47,1\%                                          \\
\bottomrule
\end{tabular}
\label{tab:generalization}
\end{center}
\end{table}

\begin{mybox}{Summary}

    \textit{Guides generated through RAR demonstrated desired intra-domain generalization by increasing the use of guide memory over new generation with stronger FM (10.2\% increase over 4 stages). Additionally, using guides from one domain (professional law) to assist in another led to an increase in aligned response generation (6.1\% for high-school psychology and 7\% for moral scenarios samples), demonstrating some degree of inter-domain generalization}
\end{mybox}  
    \section{Threats to Validity}
\label{sec:threats}

In this section, we present and discuss various threats to the validity of our study.

\subsection{Internal Validity}

One threat arises from the reliance of RAR on accurate semantic comparison between two generated outputs. When dealing with open-ended text generation, evaluating whether two samples are semantically similar is a non-trivial task. To mitigate this concern, in our evaluation we use a more constrained multiple choice Q\&A benchmark where solutions are pre-determined and already provided to the FM, leading to more expected types of output. Furthermore, we utilize vector similarity \citep{rag} and LLM-as-a-judge \citep{llm_judge} methods that have previously demonstrated successful applications in semantic comparison.

\subsection{External Validity}

While we have demonstrated the usefulness of the RAR method on a multiple-choice question-answering benchmark where input and output are reasonably constrained, such findings might not extend to other more general tasks where there are fewer constraints (e.g. creating new content rather than problem-solving). This issue can be mitigated by careful design of both guide generation and guide consumption prompts based on the desired task. Theoretically, a case-by-case solution for the majority of target applications (e.g. summarization, automation, Q\&A) can be described in a step-by-step process and thus could potentially be condensed into a step-by-step guide for use by an FM. Future work can explore applications of RAR on a wider set of different tasks.

\subsection{Construct Validity}

In our attempt to simulate real-world use of a deployed RAR method, we sequentially perform inference over all samples in the dataset and repeat it for multiple stages. However, this approach can vastly differ from how a typical user will use the system where recent sequential requests can wildly differ, and similar types of requests might not be seen for a considerable amount of time. The recurrence period of identical or highly similar requests will affect the speed at which RAR acquires new capabilities, however, it does not affect the underlying principle of the method. As such, our evaluation approach targets the ideal case so that the system can demonstrate rapid adaptation in a short amount of time.

Similarly to the above, the order in which samples are seen affects the guide learning process. This sequential dependency affects the rate at which guide memory is populated, leading to some incoming requests being able to access guides that might not have been available otherwise should the sequence of requests be changed. To mitigate this in our evaluation, we randomly shuffled the sequence in which samples are seen five times, with the results demonstrating similar behavior between permutations and reinforcing findings we describe in Section ~\ref{sec:evalaution}.
    \section{Conclusion}
\label{sec:conclusion}

In this paper, we present RAR, a new method for real-time adaptive FM routing that aims to improve static predictive routing with the goal of decreasing FM deployment costs while maintaining overall capability levels. With potential applications in cloud-edge and server environments, the proposed approach utilizes guided in-context learning to continuously improve the capabilities of weaker but cheaper FM over time and as a result reduce reliance on stronger and expensive FM. Evaluation results on the MMLU professional law dataset demonstrated a 50.2\% reduction in computational costs while maintaining 90.5\% of response quality when compared to the ideal static router, with similar trends observed with moral scenarios and high-school psychology topics of the MMLU dataset. Furthermore, experimental results demonstrated increased use of guide memory over time (10.2\% over 4 stages), pointing to intra-domain generalization of acquired guidance and leading to reduced FM inference costs.  
    
    \balance
    
    \printbibliography
\end{document}